\documentclass[letter, 12pt]{article}
\usepackage{arxiv}

\usepackage[utf8]{inputenc} 
\usepackage[T1]{fontenc}    
\usepackage{hyperref}       
\usepackage{url}            
\usepackage{booktabs}       
\usepackage{amsfonts}       
\usepackage{nicefrac}       
\usepackage{microtype}      
\usepackage{lipsum}		
\usepackage{graphicx}
\usepackage{doi}
\usepackage{caption}
\usepackage{multirow}
\usepackage{makecell}

\usepackage{mathptmx}
\usepackage[T1]{fontenc}

\title{Adversarial Adaptation\\for French Named Entity Recognition}

\author{\href{https://orcid.org/0000-0002-3416-6020}{\includegraphics[scale=0.06]{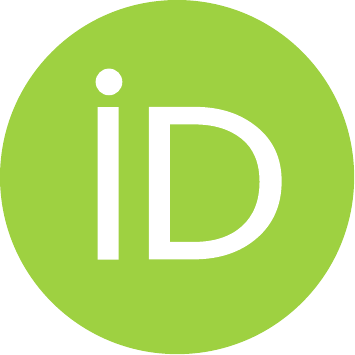}\hspace{1mm}Arjun Choudhry*}\\
	 Biometric Research Laboratory\\
  Delhi Technological University, New Delhi, India\\
  and\\
  Université du Québec à Montréal\\
	 Montréal, QC, Canada\\
  \texttt{choudhry.arjun@gmail.com} \\
  	$^*$These authors contributed equally.\\
\And 
\href{https://orcid.org/0000-0002-3462-1062}{\includegraphics[scale=0.06]{orcid.pdf}\hspace{1mm}Inder Khatri*}\\
	 Biometric Research Laboratory\\
  Delhi Technological University, New Delhi, India\\
  \texttt{inderkhatri999@gmail.com} \\	
  $^*$These authors contributed equally.\\%
\AND
\href{https://orcid.org/0000-0002-2591-324X}{\includegraphics[scale=0.06]{orcid.pdf}\hspace{1mm}Pankaj Gupta}\\
	 Biometric Research Laboratory\\
  Delhi Technological University, New Delhi, India\\
  \texttt{pankajgupta.dtu} \\
\And
\href{https://orcid.org/0000-0002-5480-5045}{\includegraphics[scale=0.06]{orcid.pdf}\hspace{1mm}Aaryan Gupta}\\
	 Biometric Research Laboratory\\
  Delhi Technological University, New Delhi, India\\
  \texttt{aryan227227@gmail.com} \\
\AND
 \href{https://orcid.org/0000-0002-1476-818X}{\includegraphics[scale=0.06]{orcid.pdf}\hspace{1mm}Maxime Nicol}\\
	 Université du Québec à Montréal\\
	 Montréal, QC, Canada\\
    \texttt{nicol.maxime@courrier.uqam.ca} \\
\AND
	\href{https://orcid.org/0000-0001-8196-2153}{\includegraphics[scale=0.06]{orcid.pdf}\hspace{1mm}Marie-Jean Meurs} \\
	Université du Québec à Montréal\\
	Montreal, QC, Canada\\
	\texttt{meurs.marie-jean@uqam.ca} \\
 \And
   \href{https://orcid.org/0000-0002-1026-0047}{\includegraphics[scale=0.06]{orcid.pdf}\hspace{1mm}Dinesh Kumar Vishwakarma}\\
    Biometric Research Laboratory\\
	Delhi Technological University\\
	New Delhi, India\\
	\texttt{dinesh@dtu.ac.in}\\
}

\date{}

\renewcommand{\shorttitle}{Adversarial Adaptation for French Named Entity Recognition}

\hypersetup{
pdftitle={Adversarial Adaptation for French Named Entity
Recognition},
pdfsubject={cs.CL},
pdfauthor={Arjun Choudhry, Inder Khatri, Pankaj Gupta, Aaryan Gupta, Maxime Nicol, Marie-Jean Meurs, Dinesh Kumar Vishwakarma},
pdfkeywords={Named Entity Recognition, French Corpora, Adversarial Adaptation, Information Retrieval, Natural Language Processing},
}
\begin{document}
\maketitle

\begin{abstract}
Named Entity Recognition (NER) is the task of identifying and classifying named entities in large-scale texts into predefined classes. 
NER in French and other relatively limited-resource languages cannot always benefit from approaches proposed for languages like English due to a dearth of large, robust datasets.
In this paper, we present our work that aims to mitigate the effects of this dearth of large, labeled datasets. 
We propose a Transformer-based NER approach for French, using adversarial adaptation to similar domain or general corpora to improve feature extraction and enable better generalization. 
Our approach allows learning better features using large-scale unlabeled corpora from the same domain or mixed domains to introduce more variations during training and reduce overfitting.
Experimental results on three labeled datasets show that our adaptation framework outperforms the corresponding non-adaptive models for various combinations of Transformer models, source datasets, and target corpora. 
We also show that adversarial adaptation to large-scale unlabeled corpora can help mitigate the performance dip incurred on using Transformer models pre-trained on smaller corpora.

\end{abstract}

\section{Introduction}
Named Entity Recognition (NER) is the task of identifying and extracting specific entities from unstructured text, and labeling them into predefined classes.
Over the years, NER models for high-resource languages, like English, have seen noticeable improvements in task performance owing to model architecture advancements and the availability of large, labeled datasets. 
In sharp contrast, languages like French still lack openly available, large-scale, labeled, robust datasets free from biases, and have few general-domain language models, and barely any domain-specific ones. 
Creating large, robust NER datasets requires manual annotation and is prohibitively expensive, calling for less labeled data-reliant approaches, particularly for limited and low-resource languages.

Over the years, a variety of deep learning-based NER approaches have been proposed~\cite{NER_Trends}. Some of the older approaches made use of external texts, or gazetteers, for the disambiguation of input words~\cite{Gazetteers_ex1}. 
Several works have used combinations of Recurrent Neural Networks like LSTMs and GRUs with Conditional Random Fields to improve model performance for a variety of NER tasks~\cite{GRU-CRF,LSTM-CRF1,LSTM-CRF2}. 

Recent works have also incorporated the use of pre-trained contextualized language models for improved NER performance. 
Copara \textit{et al.}~\cite{LM_Bio} proposed an ensemble-based NER framework for Biomedical NER in French, using various combinations of French language models. 
Liu \textit{et al.}~\cite{NER-BERT} pre-trained a NER-BERT model on a large NER corpus to counter the underlying discrepancies between the language model and the NER dataset. 
They observed significantly better performance than the standard BERT~\cite{BERT} Transformer model across nine domains. 

Researchers have also started incorporating various domain adaptation approaches for adapting NER models from high-resource domains to low-resource domains~\cite{multi-domain-NER,Cross_NER}. Peng \textit{et al.}~\cite{DA_NER} further proposed an entity-aware domain-adaptive framework using an attention layer to enable the improved transfer of features for models trained on one domain to another domain using adversarial learning. 
They observed noticeably better performance in cross-domain settings than without domain adaptation. 
However, due to the lack of robust and labeled datasets across multiple domains in limited-resource languages, these domain adaptive approaches are mostly restricted to high-resource languages. 
Wang \textit{et al.}~\cite{Astral} further introduced an adversarial perturbation approach for reduced overfitting, while also proposing the use of Gated-CNN to fuse the spatial information between adjacent words. However, they tested their approach on English datasets and noticed marginal gains over baseline approaches.

In this work, which follows our preliminary exploration~\cite{AAAI23_Arjun_arXiv}, we incorporate the use of adversarial adaptation to improve the performance of NER models in in-domain settings for French. We propose a Transformer-based NER approach, which uses adversarial adaptation to counter the lack of large-scale labeled NER datasets in French. 
This helps us evaluate the use of adversarial adaptation in enabling a model to learn improved, generalized features by adapting them to large-scale unlabeled corpora that are readily available and easy to generate for even low-resource languages. 
We further train Transformer-based NER models on labeled source datasets and use larger corpora from similar or mixed domains as target sets for improved feature learning by the model. 
Our proposed approach helps outsource wider domain and general feature knowledge from easily available large unlabeled corpora. 
We limit the purview of our evaluation to the French language in this paper. 
However, our approach could further be applied to other limited and low-resource languages for improved NER performance, as well as for other downstream tasks. 
This paper is organized as follows: the proposed methodology is introduced in Section~\ref{sec:metho}. Section~\ref{sec:expe} presents our experiments and discusses the obtained results while Section~\ref{sec:conc} concludes our findings.

\section{Proposed Methodology}
\label{sec:metho}

\subsection{Datasets and Preprocessing}
In this work, we use the WikiNER French~\cite{wikiner}, WikiNeural French~\cite{wikineural}, and Europeana French~\cite{Europeana} datasets as the labeled source datasets used for the supervised training branch in our model. The Europeana dataset contains text extracted from historic European newspapers using Optical Character Recognition (OCR). 
However, the dataset contains some OCR errors, thus making it noisy. 
This can lead to lower performance by models trained on Europeana. 

For the unlabeled target corpora, we use WikiNER and WikiNeural datasets, and the Leipzig Mixed French corpus\footnote{\url{https://wortschatz.uni-leipzig.de/en/download/French}}. 
We remove the labels for the former two datasets for use as large, unlabeled target corpora. These corpora enable us to evaluate the impact of adapting the models to similar-domain data, as well as more generalized corpora.

During preprocessing, we convert all NER tags to Inside-Outside-Beginning (IOB)~\cite{IOB} format and store the datasets in the CoNLL 2002~\cite{CoNLL} NER format for easier data input during training.


\subsection{Adversarial Adaptation to Similar Domain Corpus}
Adversarial adaptation aids in selecting domain-invariant features which are transferable between source and target datasets~\cite{Adversarial_adapt}. 
Compared to other approaches to domain adaptation, adversarial adaptation incorporates a domain discriminator into the classification framework. 
A domain discriminator acts as a domain classifier and is trained on the features retrieved by the feature extractor layer in the framework. 
It is tasked with distinguishing between the features obtained from the source and target sets. 
With the help of a Gradient Reversal Layer (GRL)~\cite{GRL}, the gradient flow of the domain discriminator is utilized to penalize the feature extractor for learning the domain-specific features, thus causing the feature extractor to learn domain-invariant features. 
GRL reverses the gradient direction and thus helps train different components of the neural network \textit{adversarially}. 
This enables the feature extractor to yield features free from domain biases present in the source domain but not the target domain.

We propose the use of adversarial adaptation of NER models to large-scale, unlabeled corpora from similar or mixed domains. 
This helps us enable the model to extract relatively more generalizable features from the same domain as the source dataset, without altering the feature extractor significantly to learn overly generalized features that reduce in-domain performance. 
Adversarial adaptation to similar domain corpora thus helps reduces the risk of overfitting on the source intricate training set-specific features, as it aids the feature extractor in extracting more generalizable features, indistinguishable from the features from the large-scale target corpora. This helps counter the dearth of large and robust training datasets in languages like French, which are readily available in English.

\begin{figure*}[t!]
     \centering
     \includegraphics[width = \textwidth]{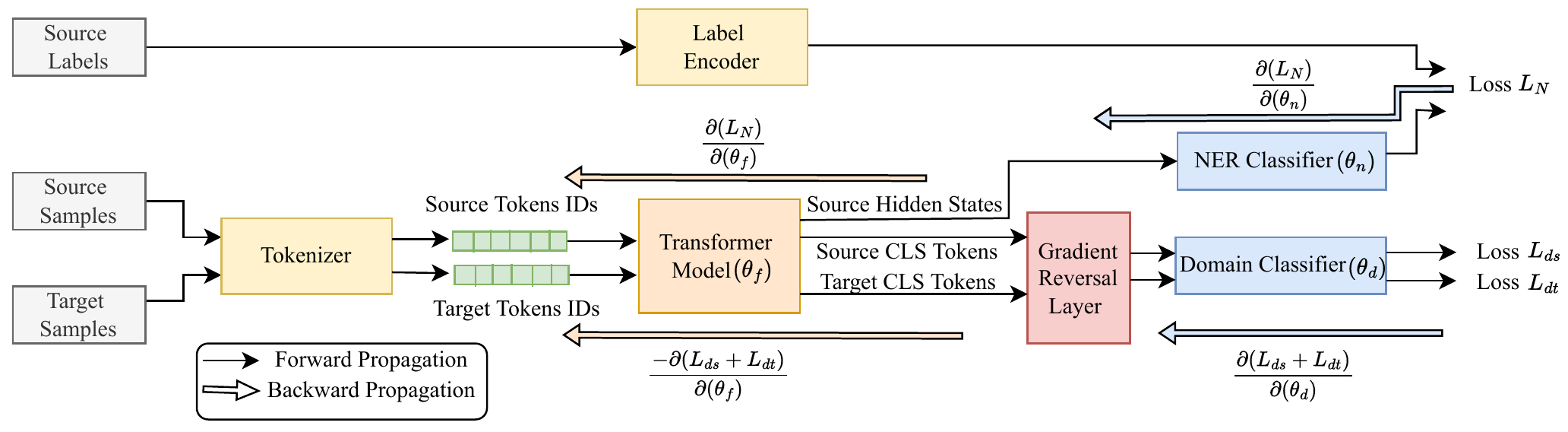}
     \hfill
   \caption{Graphical representation of our adversarial adaptation framework for training NER models on source and target sets.}
   \label{Flowchart} 
\end{figure*}

We evaluate our approach for three scenarios: source and target datasets are from the same domain; source and target datasets are from relatively different domains; and the source belongs to a certain domain while the target dataset is a mixed-domain, large-scale, general corpus. 
The latter scenario helps generalize the model further, reducing the domain-specific feature extraction in favor of features common to both the source dataset and the target corpora.
Figure~\ref{Flowchart} graphically illustrates our proposed framework. 

Our Transformer model is trained using two losses: the NER classifier loss $L_{NER}$, defined in Equation~\ref{equ:classloss}, and the adversarial loss $L_{adv}$, defined in Equation~\ref{equ:adloss}. The total loss is defined in Equation~\ref{equ:totloss}. $L_{NER}$ is the standard loss that penalizes the Transformer model's token classification error, encouraging it to make more accurate entity predictions. It is further responsible for optimizing the NER Classifier's weights. $L_{adv}$ is used adversarially, where the Transformer model is optimized in a way to maximize $L_{adv}$, while the domain classifier is optimized to minimize $L_{adv}$. To achieve this, we use a Gradient Reversal Layer between the Transformer model and the domain classifier. The Gradient Reversal Layer acts as an identity function during forward propagation, but during back-propagation, it multiplies its input by -1. This causes the back-propagated gradient to perform gradient ascent on the Transformer model with respect to the domain classifier's classification loss, rather than gradient descent.
We specifically use the adversarial domain classifier as a discriminator in our framework. 
\begin{equation}
\label{equ:classloss}
 L_{NER} \ = \ \min\limits_{\theta_{f},\theta_{n}} \sum_{i=1}^{n_{s}} L_{n}^i
\end{equation}
 
\begin{equation}
\label{equ:adloss}
 L_{adv} \ = \ \min\limits_{\theta_d} (\max\limits_{\theta_f}( \sum_{i=1}^{n_{s}} L_{ds}^i \ + \ \sum_{j=1}^{n_{t}} L_{dt}^j))
\end{equation}

\begin{equation}
\label{equ:totloss}
 L_{Total} \ = \ L_{NER} \ + \ \alpha (L_{adv})
\end{equation}

Here, $n_s$ and $n_t$ represent the number of samples in source and target sets respectively, $\theta_d$, $\theta_n$, and $\theta_f$ are the number of parameters for domain classifier, NER classifier, and Transformer model respectively, and $L_{ds}$ and $L_{dt}$ represent the Negative log-likelihood loss (NLLL) for the source and target respectively. 
We introduce another parameter $\alpha$, which is the ratio between $L_{NER}$ and $L_{adv}$ in the total loss. 
This helps us with correctly penalizing the NER classifier and the domain classifier. 
We found the optimum value of $\alpha$ to be equal to 2, as this led to the best experimental results.

\subsection{Language Models for NER}
Recent NER research has incorporated large language models due to their contextual knowledge learned during pretraining~\cite{xlnet-bilstm-crf,chinese-bert,transformer-NER}. 
We evaluate the efficacy of our proposed approach on three French language models: CamemBERT-base~\cite{camembert}, CamemBERT-Wikipedia-4GB (a variant of CamemBERT pre-trained on only 4GB of Wikipedia French corpus), and FlauBERT-base~\cite{flaubert}. 
FlauBERT is based on the BERT~\cite{BERT} architecture, while CamemBERT is based on the RoBERTa~\cite{RoBERTa} architecture. 
FlauBERT is trained on nearly half the amount of training data (71GB) as CamemBERT (138GB). 
All three language models provide 768-dimensional word embeddings. 
Comparing CamemBERT-base and CamemBERT-Wiki-4GB helps us analyze if we can replace language models pre-trained on large corpora with models pre-trained on smaller corpora adapted to unlabeled corpora during fine-tuning on a downstream task. 

\section{Experiments and Results}
\label{sec:expe}

We evaluated the performance of our approach for various combinations of language models, source datasets, and target datasets. 
Each model was evaluated on the test subset (20\% of the total data) of the source dataset. 
All domain-adaptive and baseline models were trained for up to 16 epochs. 
The training was stopped when peak validation accuracy for the in-domain validation set was achieved. 
We used a batch size of 16 for each experiment. 
We used the AdamW optimizer~\cite{AdamW}, an optimized version of Adam with weight decay with a learning rate of 0.00002 and a learning rate scheduler while training. 
We used a Transformer encoder-based layer with 512 units for feature extraction. 
Both the NER and domain classifier branches consisted of two dense layers each. 
We used a gradient reversal layer for the adversarial domain classification branch, using the last two hidden embeddings as input. These experiments are fully reproducible, and the systems are made available as open-source\footnote{\url{https://github.com/Arjun7m/AA\_NER\_Fr}} under GNU GPL v3.0.
Table~\ref{DANER-Fr_results} illustrates our results. 
Some prominent findings observed are:


\begin{table}[t!]
    \centering
    \small
    \begin{tabular}{c|c|c|ccc}
    \hline
    \hline
        \small\textbf{Model} & \small\textbf{Source} & \small\textbf{Target} & \small\textbf{Precision} & \small\textbf{Recall} & \small\textbf{F1-Score} \\\hline
        \multirow{9}{*}{\textbf{\makecell{CamemBERT-\\Wiki-4GB}}}& \multirow{3}{*}{\textbf{\makecell{WikiNER}}}& \textbf{-}&  0.911& 0.925& 0.918\\
        & & \textbf{WikiNeural}&   \textbf{0.966}&  \textbf{0.963}&  \textbf{0.969}\\
        & & \textbf{Mixed-Fr}&  0.956&  0.962&  0.959\\
    \cline{2-6}
        & \multirow{3}{*}{\textbf{\makecell{WikiNeural}}}& \textbf{-}& 0.859& 0.872& 0.866\\
        & & \textbf{WikiNER}&  \textbf{0.872}&  \textbf{0.891}&  \textbf{0.881}\\
        & & \textbf{Mixed-Fr}&  0.870 &  0.879 &  0.875\\
    \cline{2-6}
        & \multirow{3}{*}{\textbf{\makecell{Europeana}}}& \textbf{-} & 0.728 &  0.642 & 0.682\\
        & & \textbf{WikiNER}&  0.738&  \textbf{0.691}&  \textbf{0.714}\\
        & & \textbf{Mixed-Fr}&  \textbf{0.774} & 0.640 &  0.701\\
    \hline
        \multirow{9}{*}{\textbf{\makecell{CamemBERT-\\base}}}& \multirow{3}{*}{\textbf{\makecell{WikiNER}}}& \textbf{-}& 0.960& 0.968& 0.964\\
        & & \textbf{WikiNeural}&  \textbf{0.973}&  0.976&  \textbf{0.975}\\
        & & \textbf{Mixed-Fr}&  0.972&  \textbf{0.978}&  0.974\\
    \cline{2-6}
        & \multirow{3}{*}{\textbf{\makecell{WikiNeural}}}& \textbf{-}&  0.943&  0.950&  0.946\\
        & & \textbf{WikiNER}&  0.943&  \textbf{0.953} &  \textbf{0.948}\\
        & & \textbf{Mixed-Fr} &  \textbf{0.946} &  0.950 &  \textbf{0.948}\\
    \cline{2-5}
        & \multirow{3}{*}{\textbf{\makecell{Europeana}}}& \textbf{-}&  0.927 &  0.933 &  0.930\\
        & & \textbf{WikiNER}& 0.911& 0.927& 0.920\\
        & & \textbf{Mixed-Fr}&  \textbf{0.942} &  \textbf{0.943} &  \textbf{0.943}\\
    \hline
        \multirow{9}{*}{\textbf{\makecell{FlauBERT-\\base}}}& \multirow{3}{*}{\textbf{\makecell{WikiNER}}}& \textbf{-}& 0.963& 0.964& 0.963\\
        & & \textbf{WikiNeural}&  0.964&  0.968&  0.966\\
        & & \textbf{Mixed-Fr}&  \textbf{0.974}&  \textbf{0.972}&  \textbf{0.973}\\
    \cline{2-6}
        & \multirow{3}{*}{\textbf{\makecell{WikiNeural}}}& \textbf{-}& 0.934&  0.946&  0.940\\
        & & \textbf{WikiNER}&  0.935&  \textbf{0.950} &  \textbf{0.942}\\
        & & \textbf{Mixed-Fr}&  \textbf{0.941} & 0.943 &  \textbf{0.942}\\
    \cline{2-6}
        & \multirow{3}{*}{\textbf{\makecell{Europeana}}}& \textbf{-}& 0.835 &  0.863 & 0.849\\
        & & \textbf{WikiNER}&  0.855&  \textbf{0.865}&  0.860\\
        & & \textbf{Mixed-Fr}&  \textbf{0.882} & 0.854 &  \textbf{0.867}\\
    \hline
    \hline
    \end{tabular}
    \caption{Performance evaluation of our proposed adversarial adaptation approach to large-scale corpora for various combinations of models, source and target sets. We observe noticeably improved performance for the adversarial adaptation models as compared to their corresponding non-adaptive models across almost all settings.}
    \label{DANER-Fr_results}
\end{table}

\textbf{Models trained using our adversarial adaptation framework consistently outperformed their non-adaptive counterparts across all metrics.}
We observed that the adversarial adaptation models showed significant performance improvements across Precision, Recall, and F1-score over their non-adaptive counterparts across almost all combinations of source datasets, target datasets, and language models. This is beneficial for low and limited-resource languages, where adversarial adaptation to unlabelled corpora can mitigate the need for creating robust labeled datasets for training NER models.

\textbf{Adversarial adaptation can help alleviate some of the performance loss incurred on using smaller models.}
On fine-tuning the CamemBERT-Wiki-4GB model using our adversarial approach, we observed performance similar to or close to the non-adapted CamemBERT-base model for select settings. 
In fact, CamemBERT-Wiki-4GB model, when trained on the WikiNER dataset and adapted to WikiNeural corpus, outperformed the unadapted CamemBERT-base model. 
Nearly every language model reported improved results with adversarial adaptation. 
Thus, the use of adversarial adaptation during fine-tuning can act as a substitute for using larger language models for downstream tasks, thus leading to reduced computational costs for pre-training as well as fine-tuning.

\textbf{Adapting NER models to the same domain target corpus as the source dataset generally leads to slightly better performance than adapting to a mixed domain corpora.}
We observed that models adapted to a corpus from the same domain as the source dataset (like when WikiNER and WikiNeural are used as source and target datasets, or vice versa) showed similar to slightly better performance than the same models were adapted to general domain corpora. 
However, both of these cases were almost always better than the corresponding unadapted models.

\textbf{Adapting NER models to a mixed-domain target corpus generally leads to better performance than adapting the models to a corpus from a different domain corpus.}
We observed that models, when adapted to a mixed-domain corpus (like in the case of Europeana to Mixed-Fr), showed noticeably better performance than the corresponding models adapted to a corpus from a slightly different domain (as in the case of Europeana to WikiNER). 
However, in most scenarios, both of these settings led to better performance than the unadapted setting.

\section{Conclusion and Future Work}
\label{sec:conc}

In this work, we proposed a Transformer-based Named Entity Recognition framework using adversarial adaptation to large-scale similar domain or mixed domain corpora for improved feature learning in French. 
We adapted our models using our framework to corpora from the same domain as the source dataset as well as mixed-domain corpora. 
We evaluated our approach on three French language models, three target datasets, and three large-scale corpora. 
We observed noticeably improved performance using our proposed approach, as compared to models trained without our approach, across almost all Transformers and datasets. 
We further observed that adapting a model to a large-scale corpus for a downstream task can help alleviate some of the performance loss incurred by using smaller language models pre-trained on less robust corpora. 
Our proposed framework can further be applied to other low and limited-resource languages for a variety of downstream tasks. 

In the future, we intend to evaluate the efficacy of our approach for multi-lingual and cross-lingual NER, particularly for out-of-domain settings using a multi-lingual corpus as the target dataset for adaptation and multi-lingual transformer embeddings. 
This can help reduce the dependence on labeled datasets for each language for NER. 
We further aim to evaluate the impact of the underlying language script used as the target data, and how performance is affected upon using a target corpora composed of texts from languages from different scripts for improved generalization over the language-independent features.

\noindent\textbf{Acknowledgments}
This research was enabled by support provided by Calcul Québec, The Digital Research Alliance of Canada and MITACS. 

\bibliographystyle{splncs04}
\bibliography{ECIR2023/ECIR_arXiv/ECIR_arXiv_bib}

\end{document}